\documentclass[runningheads]{llncs}
\usepackage{graphicx}
\usepackage{amsmath}
\usepackage{amsfonts}
\usepackage{amssymb}
\begin{document}
\title{Generative Landmarks Guided
 Eyeglasses Removal 3D Face Reconstruction}
%
%
\author{Dapeng Zhao\inst{1} \and
Yue Qi\inst{1,2,3} }
\authorrunning{D Zhao,Y Qi.}
%
\institute{State Key Laboratory of Virtual Reality Technology and Systems,School of Computer Science and Engineering at Beihang University, Beijing, China  \\
\email{mirror1775@gmail.com}\\
 \and
 Peng Cheng Laboratory, Shenzhen, China\\
 \and
 Qingdao Research Institute of Beihang University, Qingdao, China \\
 \email{}}
\maketitle              
\begin{abstract}
    Single-view 3D face reconstruction is a fundamental Computer Vision problem of extraordinary difﬁculty. Current systems often assume the input is unobstructed faces which makes their method not suitable for in-the-wild conditions. We present a method for performing a 3D face that removes eyeglasses from a single image. Existing facial reconstruction methods fail to remove eyeglasses automatically for generating a photo-realistic 3D face "in-the-wild".The innovation of our method lies in a process for identifying the eyeglasses area robustly and remove it intelligently. In this work, we estimate the 2D face structure of the reasonable position of the eyeglasses area, which is used for the construction of 3D texture. An excellent anti-eyeglasses face reconstruction method should ensure the authenticity of the output, including the topological structure between the eyes, nose, and mouth. We achieve this via a deep learning architecture that performs direct regression of a 3DMM representation of the 3D facial geometry from a single 2D image. We also demonstrate how the related face parsing task can be incorporated into the proposed framework and help improve reconstruction quality. We conduct extensive experiments on existing 3D face reconstruction tasks as concrete examples to demonstrate the method's superior regulation ability over existing methods often break down.

\keywords{3D Face Reconstruction  \and Eyeglasses \and Occluded Scenes \and Face Parsing.}
\end{abstract}
\section{Introduction}
3D face reconstruction is an important and popular 
research field of computer vision~\cite{Blanz2003a,TuanTran2017,Gilani2018}. It is widely used in face recognition, video editing, film avatars and so on. Face 
occlusions (such as eyeglasses, respirators, eyebrow pendants and so on.) can degrade the performance of face recognition and face animation evidently. We cannot use artificial intelligence to robustly predict the 3D texture of the occluded area of the face. How to remove occlusions on face image robustly and automatically becomes one crucial problem in 3D face reconstruction processing. As the human face is one kind of particular image (the face area is not large, but there are many features, and humans are very familiar and sensitive to it), common image inpainting techniques cannot be used to remove face occlusions. The traditional image inpainting methods reconstruct the damaged image region by its same surrounding pixels, which does not consider the structure of the face. For example, if an eye of a human is occluded, the conventional inpainted face cannot reconstruct the eye image, and the output 2D face will have only one 
eye~\cite{Wang2007}. However, things have changed in recent years. Due to the rapid development of deep learning and face parsing methods, face inpainting approaches have developed rapidly. Some common extreme scenarios (\textit{i.e.}, with eyeglasses) become easy to handle.
\\3D morphable models 3DMM was proposed in 1999, which was a widely influential template reconstruction 
method~\cite{Blanz2003a,Blanz1999,Paysan2009,RN1012,RN997,RN1015}. Since the facial features are distributed very regularly, the application of the template method has continued until now. On the other hand, due to the limitation of the template's space, the expressiveness of the model is very lacking, especially the geometric details.
\\In this paper, we proposed a robust and fast face eyeglasses removal reconstruction algorithm based on face parsing and the deep learning method. \textbf{The main contributions are summarized as follows:}
\\$\bullet$\ We propose a novel algorithm that combines feature points and face parsing map to generate face which removes eyeglasses. 
\\$\bullet$\ In order to solve the problem of the invisible face area under eyeglasses occluded scenes, we propose synthesizing input face image based on Generative Adversarial Network rather than reconstructing 3D face directly. 
\\$\bullet$\ We have improved the loss function of our 3D reconstruction framework for eyeglasses occluded scenes. Our method obtains state-of-the-art qualitative performance in real-world images.  
\section{Related Work}
\subsection{Generic Face Reconstruction}Blanz \textit{et al.}~\cite{Blanz1999,Wang2020c} proposed the 3D Morphable Model (3DMM) for modeling the 3D face from a single face photo. 3DMM is a statistical model which transforms the shape and texture into a vector space representation. Though a relatively robust face model result can be achieved, the expressive power of the 3D model is limited. In addition, this method suffers from high computational costs. Rara \textit{et al.}~\cite{Rara2011} proposed a regression model between the 2D face landmarks and the corresponding 3DMM coefficient. They employed principal component regression for face model coefficient prediction. Since large facial pose changes may reduce the performance of 2D facial landmark detection, Dou \textit{et al.}~\cite{Dou2014} proposed a dictionary-based representation of 3D face shape; They then adopted sparse coding to predict model coefficients. The related comparative experiment shows that their method achieved better robustness to the previous facial landmark detection method. Following this work, Zhou \textit{et al.}~\cite{Zhou2015} also utilize a dictionary-based model; they introduced a convex formulation to estimate model parameters.
\\With the development of deep learning, 3D face reconstruction has witnessed remarkable progress in both quality and efficiency by Convolution Neural Network (CNN).
In 2017, Anh \textit{et al.}~\cite{TuanTran2017} utilized ResNet to estimate the 3D Morphable Model parameters. However, the performance of the methods is restricted due to the limitation of the 3D space defined by the face model basis or the 3DMM templates.
\subsection{Face Parsing}
\begin{figure}[htb]
    \centering
    \includegraphics[width=1.00\textwidth]{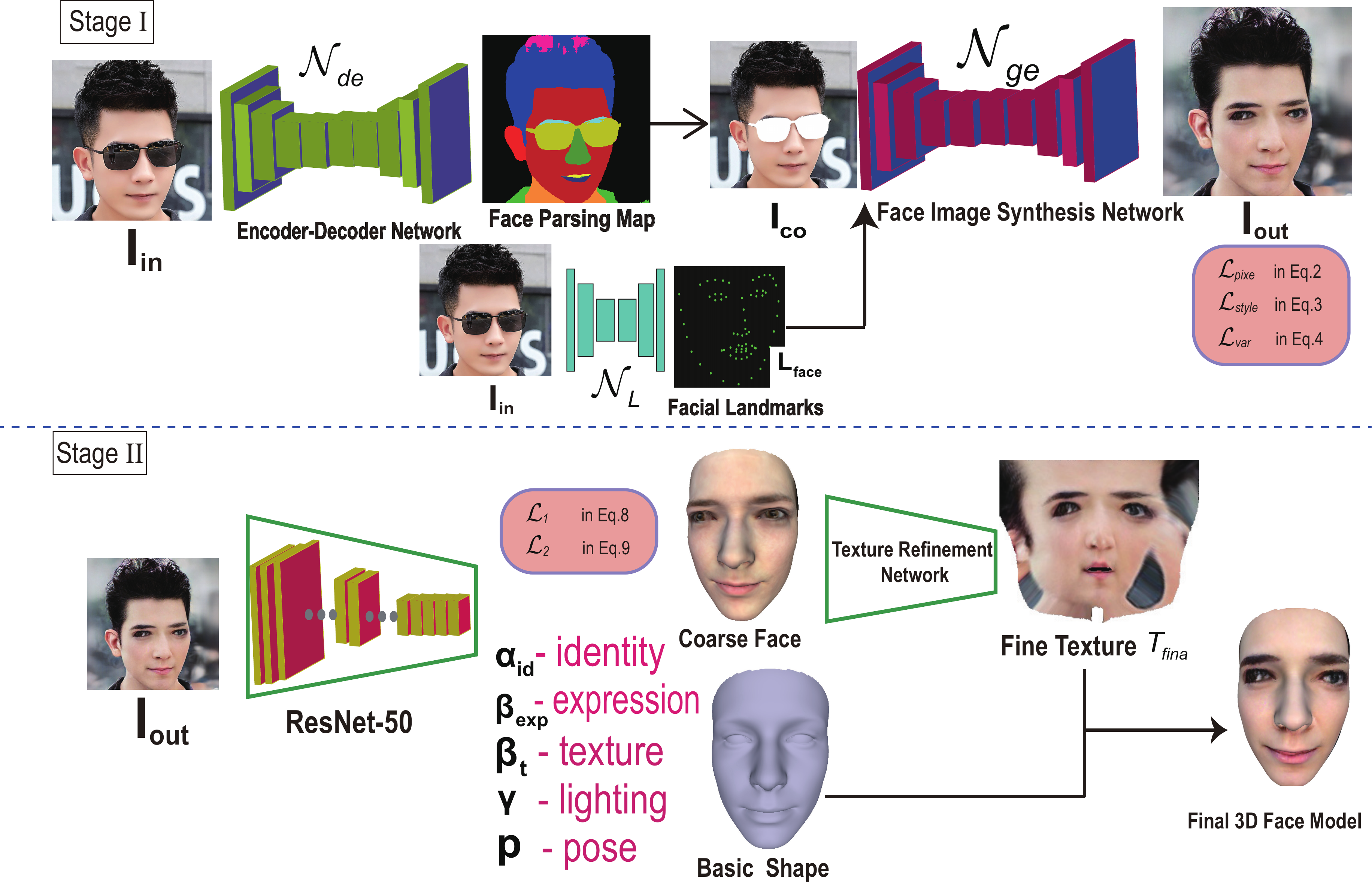}
    \caption{Our method overview. See related sections for details.} \label{fig:overall}
\end{figure}

The unique structure pattern of human face contains rich semantic representation, such as eyes, mouth, nose and so on. The low and intermediate visual features of the known region are not enough to infer the missing effective semantic features, so it is impossible to model the face 
geometry~\cite{Bertalmio2000,Huang2014}. Generate Face 
Completion~\cite{Li2017b} introduces face parsing to form regular semantic constraints. 
As shown in Figure~\ref{fig:overall}, the adoption of face parsing map can assist the face inpainting task.
\subsection{Deep Face Synthesis Methods}In the existing depth learning based face inpainting methods, due to the adoption of standard convolution layer, the synthetic pixels of the area to be inpainting comes from two parts: the valid value of the unobstructed area and the substitute value of the occluded area. This approach usually leads to color artifacts and visual blur. Deep learning has been widely used in face synthesis tasks. 
Li \textit{et al.}~\cite{Li2017b} introduced the face parsing map into the face synthesis task in order to guide GAN to generate a reasonable more brilliant face structure.
\section{Our Method}
\subsection{Landmark Estimation Network}
\begin{figure}[htb]
    \centering
    \includegraphics[width=0.50\textwidth]{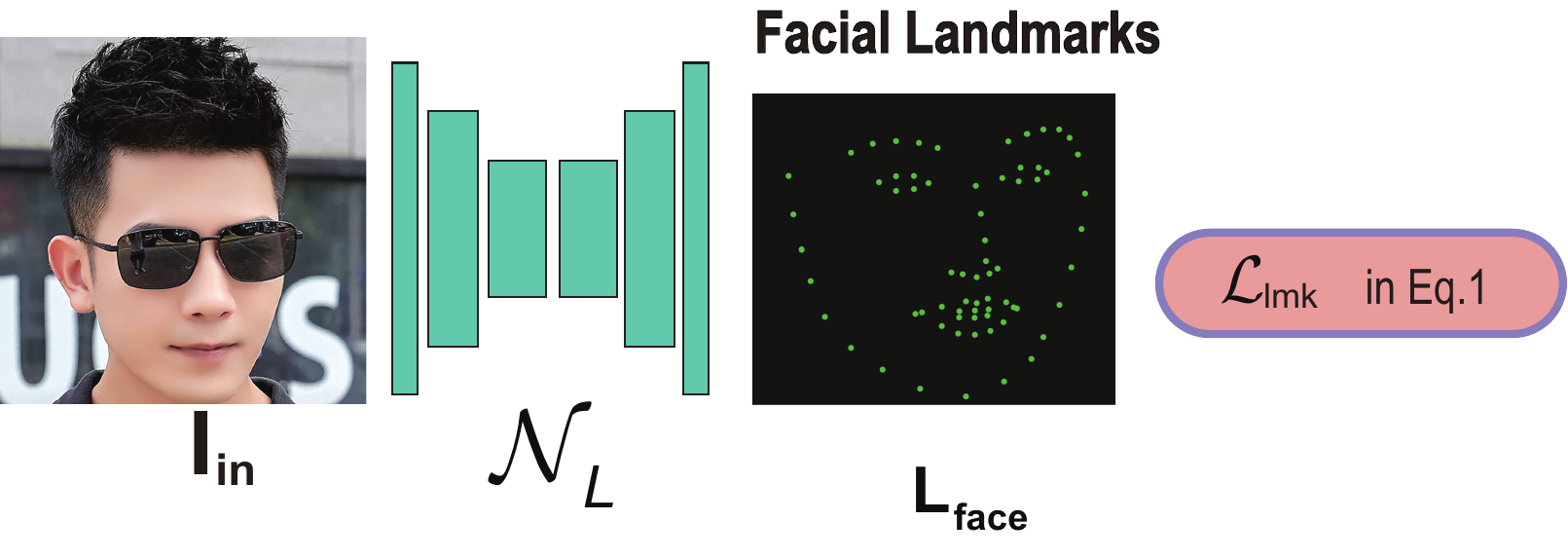}
    \caption{ Landmark prediction network of our method.} \label{fig:Landmark}
\end{figure}
In the landmark estimation task, we built the suﬃciently eﬀective landmark estimation network
 (Figure~\ref{fig:Landmark})   based on the MobileNet-V3 model~\cite{Howard2019}. In our method, accurate facial 
 landmark ${{\mathbf{L}}_{\mathbf{face}}}\in {{\mathbb{R}}^{2\times 68}}$ generation is a crucial part. The 
 network ${{\mathcal{N}}_{L}}$ aims to generate ${{\mathbf{L}}_{\mathbf{face}}}$ from a face 
 image ${{\mathbf{I}}_{\mathbf{in}}}:{{\mathbf{L}}_{\mathbf{face}}}\text{=}{{\mathcal{N}}_{L}}({{\mathbf{I}}_{\mathbf{in}}};{{\theta }_{lmk}})$, where ${{\theta }_{lmk}}$ denotes the model 
 parameters.${{\mathcal{N}}_{L}}$ is designed to extract facial features instead of face recognition, which is different from traditional 
 detectors~\cite{Kumar2018,Wu2018a}. We set the loss function as follows:
\begin{equation}
    {{\mathcal{L}}_{lmk}}=\left\| {{\mathbf{L}}_{\mathbf{face}}}-{{\mathbf{L}}_{\mathbf{gt}}} \right\|_{2}^{2}
 \end{equation}
 where ${{\mathbf{L}}_{\mathbf{gt}}}$ denotes the ground truth face landmarks 
 and ${{\left\| \cdot  \right\|}_{2}}$ denotes the ${{L}_{2}}$ norm.
\subsection{Face Synthesis Module}
\begin{figure}[htb]
    \centering
    \includegraphics[width=0.50\textwidth]{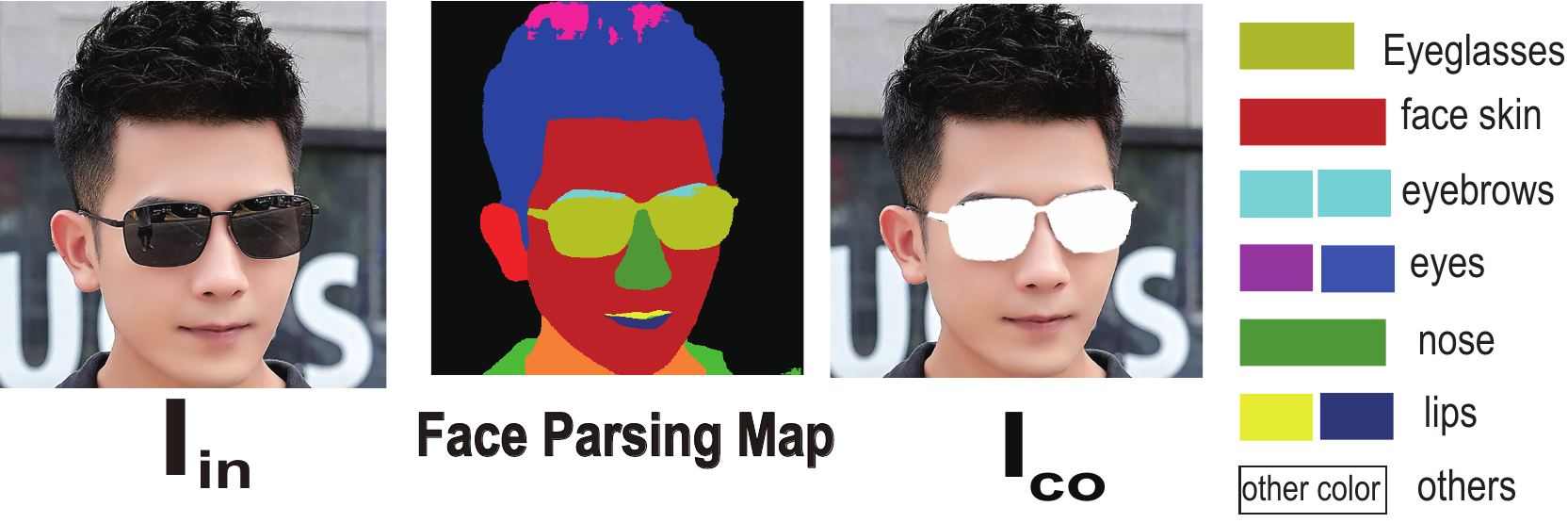}
    \caption{Face parsing module of our method.} \label{fig:FaceSynthesis}
\end{figure}
Overall, We design the synthesis 
module ${{\mathcal{N}}_{s}}$ to synthesize a $2$D image of a human face without 
eyeglasses. The module ${{\mathcal{N}}_{s}}$ consists of three parts: \textit{deleter, generator} and \textit{discriminator.}
\subsubsection{{Deleter.}} Normally, the task of the deleter   is to delete the occluded eyeglasses areas ${{\mathbf{I}}_{\mathbf{m}}}$ of the facial 
features in the input image ${{\mathbf{I}}_{\mathbf{in}}}$ (Figure \ref{fig:FaceSynthesis}). 
Overall, the deleter ${{\mathcal{N}}_{de}}$ is based on the U-Net structure~\cite{Ronneberger2015}. Inspired by the annotated face dataset 
CelebAMask-HQ~\cite{Lee2020}, we used the encoder-decoder 
architecture ${{\mathcal{N}}_{de}}$ to estimate pixel-level label classes.
Given the input face image ${{\mathbf{I}}_{\mathbf{in}}}\in {{\mathbb{R}}^{\text{H}\times \text{W}\times \text{3}}}$, we applied the trained 
model ${{\mathcal{N}}_{de}}$ to obtain the face parsing map $\mathbf{M}\in {{\mathbb{R}}^{\text{H}\times \text{W}\times 1}}$.
According to the map $\mathbf{M}$, we identify and delete the eyeglasses area ${{\mathbf{I}}_{\mathbf{m}}}$  to obtain the corrupted 
image ${{\mathbf{I}}_{\mathbf{co}}}$.
\subsubsection{Generator.}The 
generator ${{\mathcal{N}}_{ge}}$ is also based on the U-Net structure, which desires to synthesize the full face by taking corrupted 
images ${{\mathbf{I}}_{\mathbf{co}}}$ and landmarks $\mathbf{L}$ (${{\mathbf{L}}_{\mathbf{face}}}$ or ${{\mathbf{L}}_{\mathbf{gt}}}$). The generator can be 
formulated as ${{\text{I}}_{\mathbf{out}}}={{\mathcal{N}}_{ge}}({{\mathbf{I}}_{\mathbf{co}}},L;{{\theta }_{ge}})$, 
with ${{\theta }_{ge}}$  the trainable parameters.
\subsubsection{Discriminator.} The purpose of the discriminator is to judge whether the data distribution meets our requirements. The ambition of face synthesis is achieved when the generated results are not distinguishable from the real ones.
\\\textit{Loss of Discriminator.}We use a combination of an adversarial loss, a per-pixel loss, a perceptual loss, a style loss, a total variation loss and an adversarial loss, for training the face synthesis network.
\\The per-pixel loss is formulated as follows:
\begin{equation}
    {{\mathcal{L}}_{pixe}}=\frac{1}{S}\left\| {{\mathbf{I}}_{\mathbf{out}}}-{{\mathbf{I}}_{\mathbf{in}}} \right\|
\end{equation}
where $S$ denotes the mask size and $\left\| \cdot  \right\|$ denotes 
the ${{L}_{1}}$ norm. Here, $S$ is the denominator and its role is to adjust the penalty. A straightforward objective of per-pixel loss is to minimize the differences between the input face images and the synthetic images. It should be pointed out that our input image will not contain occlusion, so we don't need to consider this.
\\The style loss computes the style distance between two images as follows:
\begin{equation}
    {{\mathcal{L}}_{style}}=\sum\limits_{\text{n}}{\frac{1}{{{O}_{n}}\times {{O}_{n}}}\left\| \frac{{{G}_{n}}({{\mathbf{I}}_{\mathbf{out}}}\odot {{\mathbf{I}}_{\mathbf{m}}})-{{G}_{n}}({{\mathbf{I}}_{\mathbf{in}}}\odot {{\mathbf{I}}_{\mathbf{m}}})}{{{O}_{n}}\times {{H}_{n}}\times {{W}_{n}}} \right\|}
\end{equation}
where ${{G}_{\text{n}}}\text{(x)=}{{\varphi }_{n}}{{(x)}^{T}}{{\varphi }_{n}}(x)$ denotes the Gram Matrix 
corresponding to ${{\varphi }_{n}}(x)$,${{\varphi }_{n}}(\cdot )$ denotes the ${{O}_{n}}$ feature maps with the size ${{H}_{n}}\times {{W}_{n}}$ of the $n$-th layer.
\\Due to the use of the normalization tool, the synthesized face may have artifacts, checkerboards, or water droplets. We define the total variation loss as:
\begin{equation}
    {{\mathcal{L}}_{\operatorname{var}}}=\frac{1}{{{\text{P}}_{{{\mathbf{I}}_{\mathbf{in}}}}}}\left\| \nabla {{\mathbf{I}}_{\mathbf{out}}} \right\|
\end{equation}
where ${{\text{P}}_{{{\mathbf{I}}_{\mathbf{in}}}}}$ is the pixel number of ${{\mathbf{I}}_{\mathbf{in}}}$ and $\nabla$ is the first order derivative, containing ${{\nabla }_{h}}$ (horizontal) and ${{\nabla }_{v}}$ (vertical).
\\The total loss with respect to the face synthesis module:
\begin{equation}
    {{\mathcal{L}}_{fsm}}={{\lambda }_{pixe}}{{\mathcal{L}}_{pixe}}+{{\lambda }_{style}}{{\mathcal{L}}_{style}}+{{\lambda }_{\operatorname{var}}}{{\mathcal{L}}_{\operatorname{var}}}
\end{equation}
Here, we use ${{\lambda }_{pixe}}=1$, ${{\lambda }_{style}}=250$  and ${{\lambda }_{\operatorname{var}}}=0.1$ in our experiments.
\subsection{3D Face Reconstruction}
The classic single-view 3D face reconstruction methods utilize a 
3D template model (\textit{e.g.}, 3DMM) ) to fit the input face image~\cite{Blanz1999,Paysan2009}. This type of method usually consists of two steps: face alignment and regressing the 3DMM coefficients. The seminal 
work~\cite{Blanz2003a,Gerig2018,RN1016} describe the 3D face space with PCA:
\begin{equation}
    \mathbf{S}=\overline{\mathbf{S}}+{{\mathbf{A}}_{\mathbf{id}}}{{\boldsymbol{\alpha }}_{\mathbf{id}}}+{{\mathbf{B}}_{\mathbf{exp}}}{{\boldsymbol{\beta }}_{\mathbf{exp}}},\mathbf{T}=\overline{\mathbf{T}}+{{\mathbf{B}}_{\mathbf{t}}}{{\boldsymbol{\beta }}_{\mathbf{t}}}
\end{equation}
where $\overline{\mathbf{S}}$ and $\overline{\mathbf{T}}$ denote the mean shape and texture, ${{\mathbf{A}}_{\mathbf{id}}}$,${{\mathbf{B}}_{\mathbf{exp}}}$ and ${{\mathbf{B}}_{\mathbf{t}}}$ denote the PCA bases of identity, expression and texture.
${{\mathbf{\alpha }}_{\mathbf{id}}}\in {{\mathbb{R}}^{80}}$ and ${{\mathbf{\beta }}_{\mathbf{exp}}}\in {{\mathbb{R}}^{64}}$, and ${{\mathbf{\beta }}_{\mathbf{t}}}\in {{\mathbb{R}}^{80}}$ are the corresponding 3DMM coefficient vectors.
\\After the 3D face is reconstructed, it can be projected onto the image plane with the perspective projection:
\begin{equation}
    {{\mathbf{V}}_{\mathbf{2d}}}\left( \mathbf{P} \right)=f*{{\mathbf{P}}_{\mathbf{r}}}*\mathbf{R}*{{\mathbf{S}}_{\mathbf{mod}}}+{{\mathbf{t}}_{\mathbf{2d}}}
\end{equation}
where ${{V}_{2d}}\left( \mathbf{P} \right)$ denotes the projection function that turned the 3D model into 2D face positions,$f$ denotes the scale factor,${{\mathbf{P}}_{\mathbf{r}}}$ denotes the projection matrix, $\mathbf{R}\in SO(3)$ denotes the rotation matrix and ${{\mathbf{t}}_{\mathbf{2d}}}\in {{\mathbb{R}}^{3}}$ denotes the translation vector.
\\Therefore, we approximated the scene illumination with Spherical Harmonics (SH)~\cite{Ramamoorthi2001,Ramamoorthi2001a,Mueller2006,RN1017} parameterized 
by coefﬁcient vector $\gamma \in {{\mathbb{R}}^{9}}$. In summary, the unknown parameters to be learned can be denoted 
by a vector $y=({{\boldsymbol{\alpha }}_{\mathbf{id}}},{{\boldsymbol{\beta }}_{\mathbf{exp}}},{{\boldsymbol{\beta }}_{\mathbf{t}}},\boldsymbol{\gamma },\mathbf{p})\in {{\mathbb{R}}^{239}}$, 
where $\mathbf{p}\in {{\mathbb{R}}^{6}}=\{\mathbf{pitch},\mathbf{yaw},\mathbf{roll},f,{{\mathbf{t}}_{\mathbf{2D}}}\}$ denotes face poses. In this work, we used a fixed 
ResNet-50~\cite{He2016} network to regress these coefficients. 
\\The corresponding loss function consists of two parts:pixel-wise loss and face feature loss.
\subsubsection{Pixel-wise Loss.}The purpose of this loss function is very simple, which is to minimize the difference between the input image $\mathbf{I}_{\mathbf{out}}^{(i)}$ and the rendered image $\mathbf{I}_{\mathbf{y}}^{(i)}$. The rendering layer renders back an image $\mathbf{I}_{\mathbf{y}}^{(i)}$  to compare with the image $\mathbf{I}_{\mathbf{out}}^{(i)}$ . The pixel-wise loss is formulated as:
\begin{equation}
    {{\mathcal{L}}_{1}}={{\left\| \mathbf{I}_{\mathbf{out}}^{(i)}-\mathbf{I}_{\mathbf{y}}^{(i)} \right\|}_{2}}
\end{equation}
where $i$ denotes pixel index and ${{\left\| \cdot  \right\|}_{2}}$ denotes the ${{L}_{2}}$ norm.
\subsubsection{Face Features Loss.}We introduce a loss function at the face recognition level to reduce the difference between the 3D model of the face and the 2D image. The loss function computes the feature diﬀerence between the input image $\mathbf{I}_{\mathbf{out}}^{{}}$ and rendered image $\mathbf{I}_{\mathbf{y}}^{{}}$ . We deﬁne the loss as a cosine distance:
\begin{equation}
    {{\mathcal{L}}_{2}}=1-\frac{<G(\mathbf{I}_{\mathbf{out}}^{{}}),G(\mathbf{I}_{\mathbf{y}}^{{}})>}{\left\| G(\mathbf{I}_{\mathbf{out}}^{{}}) \right\|\cdot \left\| G(\mathbf{I}_{\mathbf{y}}^{{}} \right\|}
\end{equation}
where ${G}(\cdot )$  denotes the feature extraction function by FaceNet~\cite{Schroff2015}, $<\cdot ,\cdot >$ denotes the inner product.
\\In summary, we used the loss function ${{\mathcal{L}}_{3D}}$ to reconstruct the basic shape of the face. We set ${{\mathcal{L}}_{3D}}={{\lambda }_{1}}{{\mathcal{L}}_{1}}+{{\lambda }_{2}}{{\mathcal{L}}_{2}}$, where ${{\lambda }_{1}}{=}1.4$  and  ${{\lambda }_{2}}{=0}{.25}$ respectively in all our experiments. We then used a coarse-to-fine graph convolutional network based on the frameworks of Lin \textit{et al.}~\cite{Lin2020} for producing the fine texture ${{T}_{fina}}$ .
\section{Experimental Details and Results}
\subsection{Implementation Details}
In consideration of the module 
of ${{\mathcal{N}}_{cont}}$,we used the ground truth of CelebA-HQ datasets~\cite{Karras2017} as the reference.Considering the 
generator ${{\mathcal{N}}_{ge}}$ ,it consists of three gradually down-sampled encoding blocks, followed by seven residual blocks with dilated convolutions and a long-short term attention block. Then, the decoder processes the feature maps gradually up-sampled to the same size as input.
\subsection{Qualitative Comparisons with Recent Arts}
\begin{figure}[htb]
    \centering
    \includegraphics[width=0.80\textwidth]{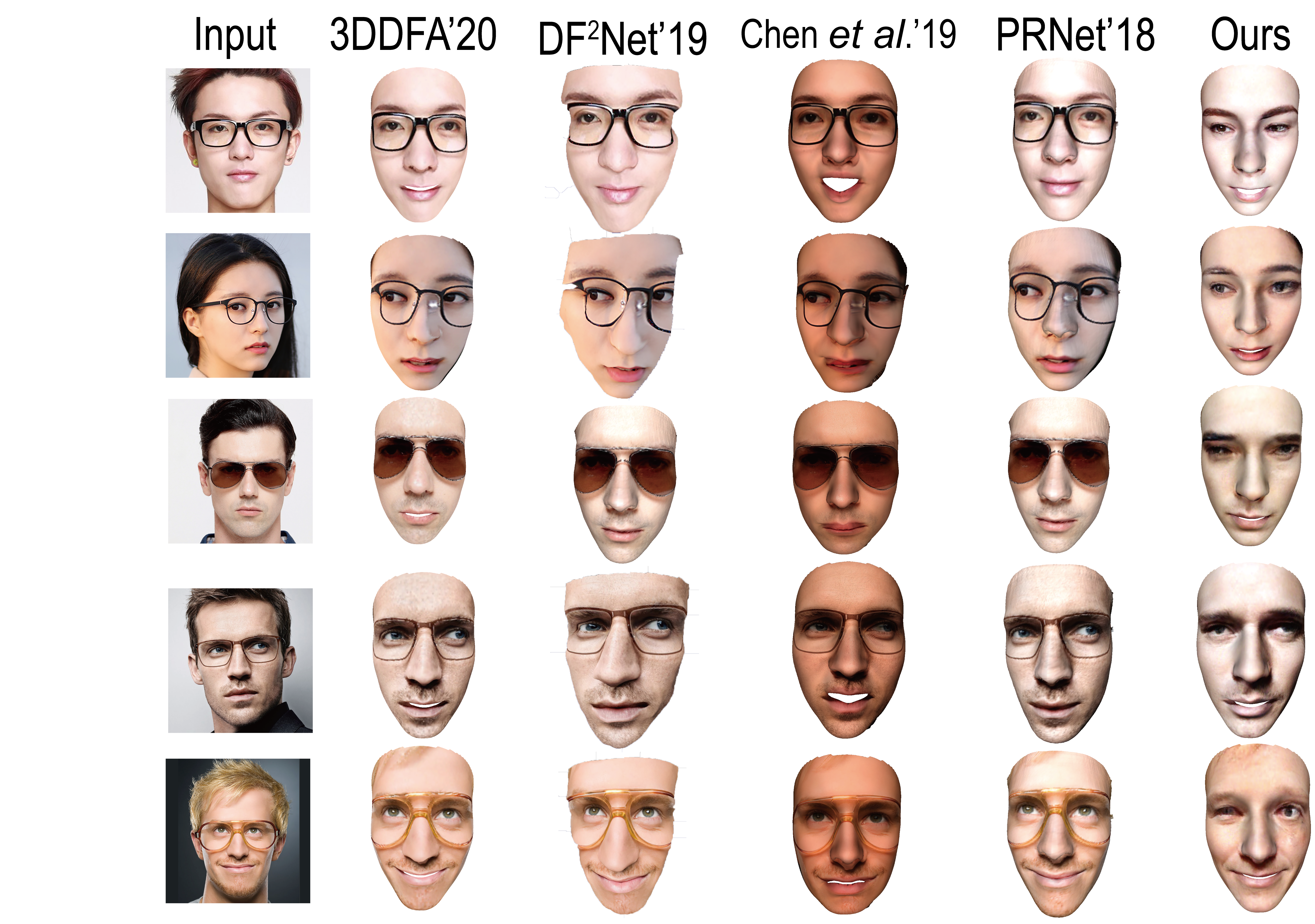}
    \caption{Comparison of qualitative results. Baseline methods from left to right: 3DDFA, DF2Net, Chen \textit{et al.},PRNet, and our method.} \label{fig:dingxingResult}
\end{figure}
Figure \ref{fig:dingxingResult} shows our experimental results compared with the 
others~\cite{Guo2020b,Zeng2019,Chen2019,Feng2018}. The result shows that our method is far superior to other frameworks. Our 3D reconstruction method can handle eyeglasses occluded scenes, such as transparent glasses and sunglasses. Other frameworks can not handle eyeglasses well; they are more focused on the generation of high-definition textures.
\subsection{Quantitative Comparison}
\subsubsection{Comparison result on the MICC Florence datasets}
\begin{figure}[htb]
    \centering
    \includegraphics[width=0.60\textwidth]{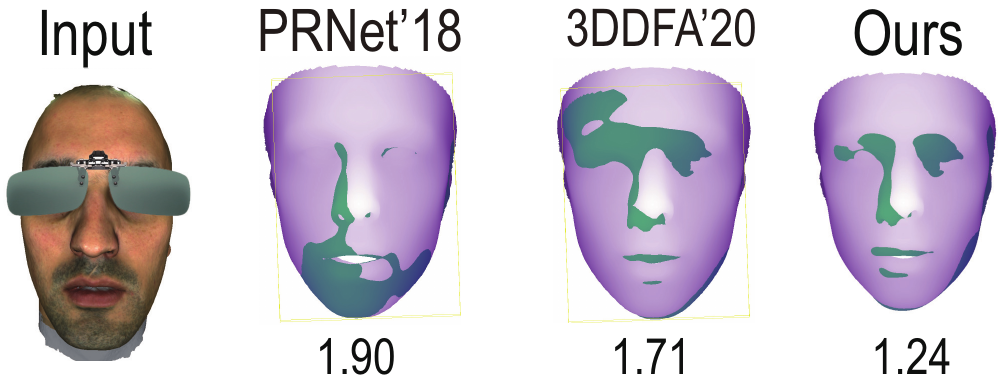}
    \caption{Comparison of error heat maps on the MICC Florence datasets. Digits denote $90$\% error (mm).} \label{fig:MICC}
\end{figure}
MICC Florence dataset~\cite{Bagdanov2011} is a 3D face dataset that contains $53$ faces with their ground truth models. We artificially added eyeglasses as input. We calculated the average $90$\% largest error between the generative model and the ground truth model.
Figure~\ref{fig:MICC} shows that our method can effectively handle eyeglasses.
\subsubsection{Eyeglasses invariance of the foundation shape}
\begin{figure}[htb]
    \centering
    \includegraphics[width=0.80\textwidth]{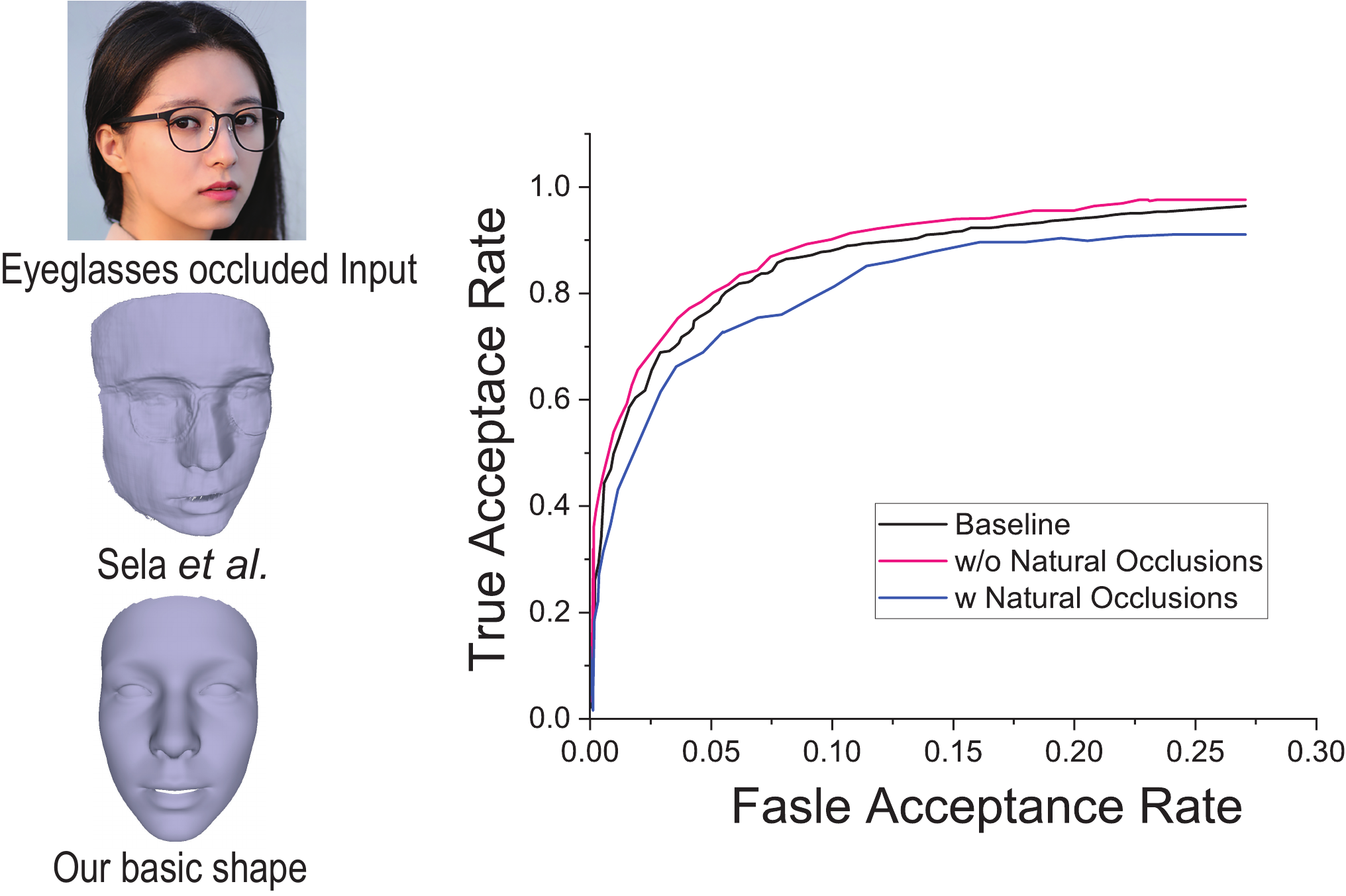}
    \caption{Reconstructions with eyeglasses. Left: Qualitative results of Sela \textit{et al.}~\cite{Isola2017} and our shape. Right: LFW veriﬁcation ROC for the shapes, with and without eyeglasses.} \label{fig:LFW}
\end{figure}
Our choice of using the ResNet-50 to regress the shape coefficients is motivated by the unique robustness to extreme viewing conditions in the paper of 
Deng \textit{et al.}~\cite{Richardson2016}.To fully support the application of our method to occluded face images, we test our system on the Labeled Faces in the Wild 
datasets (LFW)~\cite{Pan2019} . We used the same face test system 
from Anh \textit{et al.}~\cite{Extreme2018}, and we refer to that paper for more details.
\\Figure~\ref{fig:LFW} (left) shows the sensitivity of the method of 
Sela \textit{et al.}~\cite{Sela2017}. Their result clearly shows the outline of a finger. Their failure may be due to more focus on local details, which weakly regularizes the global shape. However, our method recognizes and regenerates the occluded area. Our method much robust provides a natural face shape under eyeglasses scenes. Though 3DMM also limits the details of shape, we use it only as a foundation and add refined texture separately.
\begin{table}
    \centering
    \caption{Quantitative evaluations on LFW.}
    \label{biaoge:01}
    \begin{tabular}{p{0.9in}c c c c}
    \hline
    \multicolumn{1}{l}{Method} & \multicolumn{1}{l}{100\%-EER} & \multicolumn{1}{l}{Accuracy} & nAUC       \\ \hline
    Tran \textit{et al.}~\cite{TuanTran2017}\qquad  & $89.40\pm1.52$\qquad\qquad                    & $89.36\pm1.25$\qquad\qquad                   & $95.90\pm0.95$ \\
    Ours (w/ Gla)\qquad                            & $84.77\pm1.23$\qquad\qquad                    & $87.05\pm0.89$\qquad\qquad                   & $92.77\pm1.26$ \\
    Ours (w/o Gla)\qquad                           & $89.33\pm1.15$\qquad\qquad                    & $89.80\pm0.89$\qquad\qquad                   & $96.09\pm0.61$ \\ \hline
    \end{tabular}
\end{table}
\\We further quantitatively verify the robustness of our method to eyeglasses. Table 1 (top) reports veriﬁcation results on the LFW benchmark with and without eyeglasses (see also ROC in Figure~\ref{fig:LFW}-right). Though eyeglasses clearly impact recognition, this drop of the curve is limited, demonstrating the robustness of our method.
\section{Conclusions}
We propose a novel method to reconstruct a 3D face model from an eyeglass occluded RGB face photo. Given the input image and a pre-trained ResNet, we fit the face model to a template model (3DMM). In order to robustly reconstruct RGB face without glasses, we design a deep learning network, which remakes reasonable texture intelligently. Comprehensive experiments have shown that our method outperforms previous arts by a large margin in terms of both accuracy and robustness.
\section{Acknowledgment}
This paper is supported by National Natural Science Foundation of China (No. 62072020), National Key Research and Development Program of China (No. 2017YFB1002602), Key-Area Research and Development Program of Guangdong Province (No. 2019B010150001) and the Leading Talents in Innovation and Entrepreneurship of Qingdao (19-3-2-21-zhc).

%
%
%
\bibliographystyle{splncs04}
\bibliography{MMM2022.bib}

\end{document}